\documentclass[conference]{IEEEtran}
\pdfoutput=1
\IEEEoverridecommandlockouts
\usepackage{cite}
\usepackage{amsmath,amssymb,amsfonts}
\usepackage{algorithmic}
\usepackage{graphicx}
\usepackage{textcomp}
\usepackage{xcolor}
\usepackage{subfig}
\usepackage{url}
\def\BibTeX{{\rm B\kern-.05em{\sc i\kern-.025em b}\kern-.08em
    T\kern-.1667em\lower.7ex\hbox{E}\kern-.125emX}}

\newcommand{\norm}[1]{\left\lVert#1\right\rVert}

\begin{document}

\title{On the biological plausibility of orthogonal initialisation for solving gradient instability in deep neural networks}

\author{\IEEEauthorblockN{1\textsuperscript{st} Nikolay Manchev}
\IEEEauthorblockA{\textit{Department of Informatics} \\
\textit{King’s College London}\\
London, WC2R 2LS \\
nikolay.manchev@kcl.ac.uk}
\and
\IEEEauthorblockN{2\textsuperscript{nd} Michael Spratling}
\IEEEauthorblockA{\textit{Department of Informatics} \\
\textit{King’s College London}\\
London, WC2R 2LS \\
michael.spratling@kcl.ac.uk}
}

\maketitle

\begin{abstract}
Initialising the synaptic weights of artificial neural networks (ANNs) with orthogonal matrices is known to alleviate vanishing and exploding gradient problems. A major objection against such initialisation schemes is that they are deemed biologically implausible as they mandate factorization techniques that are difficult to attribute to a neurobiological process. This paper presents two initialisation schemes that allow a network to naturally evolve its weights to form orthogonal matrices, provides theoretical analysis that pre-training orthogonalisation always converges, and empirically confirms that the proposed schemes outperform randomly initialised recurrent and feedforward networks.
\end{abstract}

\begin{IEEEkeywords}
neural networks, initialisation, orthogonality.
\end{IEEEkeywords}

\section{Introduction}
The vanishing gradient problem is a difficulty associated with training ANNs using gradient descent optimisation, more specifically -- backpropagation. The problem has been thoroughly investigated by Bengio et al.~\cite{Bengio:1994:LLD:2325857.2328340} and was initially demonstrated for recurrent neural networks. However, this stability problem is generally agnostic to the network architecture. Glorot \& Bengio~\cite{Glorot10understandingthe} show analytically that deep feedforward networks are also susceptible to vanishing gradients, and similar analysis has been performed for convolutional neural networks (CNNs) with ReLU activation functions~\cite{DBLP:journals/corr/HeZR015}. 

\subsection{Vanishing gradient in recurrent networks}\label{sec:vanishing_rnn}
\begin{figure}[tbp]
\begin{center}
\subfloat[\label{sub:1}]{\includegraphics[height=.9in]{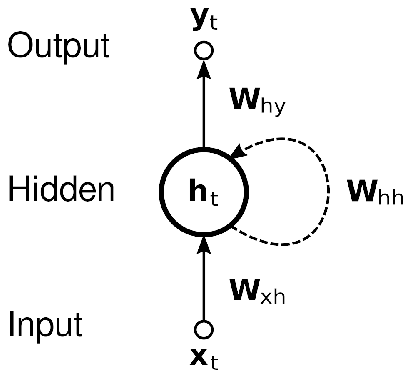}}
\hspace*{\fill}
\subfloat[\label{sub:2}]{\includegraphics[height=.9in]{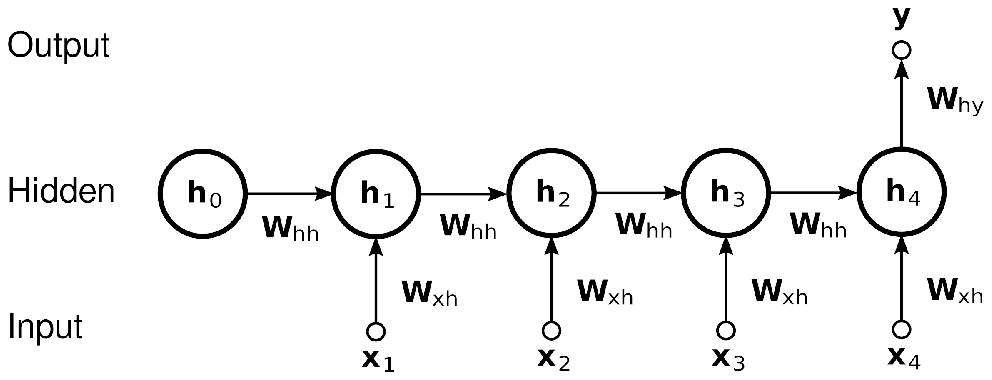}}
\end{center}
\caption{\textbf{(a)} A Simple Recurrent Network with one input, one output, and one recurrent hidden unit \textbf{(b)} An SRNN ``unrolled'' for four time steps ($t \in [1,4]$). 
}
\label{fig:SRN-BPTT}
\end{figure}

Figure~\ref{sub:1} shows a simple recurrent neural network (SRNN) architecture. The network receives an input $\textbf{x}_t$ at time $t$ and produces an output $\textbf{y}_t$ using a hidden unit $\textbf{h}_t$. The hidden unit has a recurrent connection to itself, mapping the input to some context information acquired in previous time steps. The values of the hidden unit are given by

\begin{equation}\label{eq:ht}
	\textbf{h}_t = \sigma (\textbf{W}_{xh} \cdot \textbf{x}_t + \textbf{W}_{hh} \cdot \textbf{h}_{t-1} + \textbf{b}_h)
\end{equation}

\noindent where $\textbf{W}_{xh}$ is a matrix of synaptic weights between the input and the hidden layer, $\textbf{W}_{hh}$ is a matrix of synaptic weights for the hidden layer (transition matrix), $\textbf{b}_{h}$ is a bias term, and $\sigma(\cdot)$ is a non-linear activation function. For classification problems the output of the network is given by

\begin{equation}\label{eq:softmax}
	\textbf{y}_t = \text{softmax} (\textbf{W}_{hy} \cdot \textbf{h}_t + \textbf{b}_y)
\end{equation}

\noindent where $\textbf{W}_{hy}$ is a matrix of weights between the hidden and output layers, $\textbf{b}_{y}$ is a bias term, and $\text{softmax}(\cdot)$ is the normalised exponent function. 

The weight matrices are usually initialised at random and the biases are set to vectors of zeros. A form of supervised learning is typically used to adjust these parameters. A common gradient-based technique for finding optimal parameters is \textit{backpropagation through time} or BPTT \cite{werbos1990backpropagation}. BPTT operates by ``unrolling'' the network over a time interval $[t_1, t_\text{max}]$ and treating it as a simple feedforward network (Figure~\ref{sub:2}). An error $\mathcal{E}$ is defined at time step $t_{\text{max}}$ and is used to measure the distance between the output \textbf{y} and some desired target \textbf{t}.

In the training phase all network parameters \textbf{W} (weight matrices and biases) can be adjusted by computing the gradient of $\mathcal{E}$ w.r.t \textbf{W} and by subtracting a ratio of the gradients, governed by the learning parameter $\alpha$, from the weights:

\begin{equation}\label{eq:update}
	\textbf{W}_{t+1} := \textbf{W}_{t} - \alpha  \frac{\partial \mathcal{E}}{\partial \textbf{W}_t}
\end{equation}

The process of presenting inputs, computing the error, and adjusting $\textbf{W}$ is repeated until certain convergence criterion is satisfied. The vanishing gradient problem describes a situation when the gradient becomes vanishingly small, thus preventing Equation \ref{eq:update} from efficiently updating the weights. This can be shown analytically for the gradients of the transition matrix. The derivative of $\mathcal{E}$ with respect to $\textbf{W}_{hh}$ can be expressed as a sum of the derivatives at each $t \in [1, t_{\text{max}}]$.

\begin{equation}
	\frac{\partial \mathcal{E}}{\partial \textbf{W}_{hh}} = \sum\limits_{t = 1}^{t_{\text{max}}} \frac{\partial \mathcal{E}}{\partial \boldsymbol{y}} \frac{\partial \boldsymbol{y}}{\partial \textbf{h}_{t_{\text{max}}}} \frac{\partial \textbf{h}_{t_{\text{max}}}}{\partial \textbf{h}_t} \frac{\partial \textbf{h}_t}{\partial \textbf{W}_{hh}}
\end{equation}

The $\frac{\partial \textbf{h}_{t_{\text{max}}}}{\partial \textbf{h}_t}$ term, which transports the error from step $t_{\text{max}}$ back to step $t$, is a product of Jacobian matrices such that

\begin{equation*}
	\frac{\partial \textbf{h}_{t_{\text{max}}}}{\partial \textbf{h}_t} = \prod \limits_{i = t + 1}^{t_{\text{max}}} \frac{\partial \textbf{h}_i}{\partial \textbf{h}_{i-1}} = \prod \limits_{i = t + 1}^{t_{\text{max}}} \textbf{W}^T_{hh} \text{diag} [\sigma(\textbf{h}_{i-1})]
\end{equation*}

As $t_{\text{max}}$ grows, this product of Jacobian matrices can exponentially decrease to zero. It has been shown that it is sufficient for the spectral radius (operator 2-norm) of $\textbf{W}_{hh}$ to be smaller than 1 for the gradients to vanish \cite{Pascanu2012}.

\subsection{Vanishing gradient in feedforward networks}\label{sec:vanishing-ffnet}

The vanishing gradient problem is not limited to recurrent networks. The same difficulties apply to deep feedforward networks. Consider a feedforward network with input $\boldsymbol{x}$ and hidden layers $\textbf{h}_\ell$ where $\ell \in [1, \ell_\text{max}]$ that computes a function

\begin{equation}
	\Phi(\boldsymbol{x}) = ( \textbf{h}_{\ell_\text{max}} \circ \textbf{h}_{\ell_\text{max-1}} \circ \dots \circ \textbf{h}_1 ) (\boldsymbol{x})
\end{equation}

\noindent where the output of the hidden layers and the global output is given by

\begin{flalign}\label{eq:ff-define}
        & \boldsymbol{v}_1 = \textbf{W}_1 \cdot {x} + \textbf{b}_1 \notag \\
		& \boldsymbol{v}_\ell = \textbf{W}_\ell \cdot \textbf{h}_{\ell-1} + \textbf{b}_\ell \text{, for } \ell \in [2, \ell_\text{max}] \\
		& \textbf{h}_\ell = \sigma ( \boldsymbol{v}_\ell )  \notag \\
   	   & \textbf{y} = \text{softmax} (\textbf{h}_{\ell_\text{max}} \cdot \textbf{W}_{\ell_\text{max+1}} + \textbf{b}_y) \notag
\end{flalign}






One way to confirm that deep feedforward networks are susceptible to the vanishing gradient problem is to consider the SRNN to be the deepest possible form of a feedforward network and claim that the same reasoning applies \cite{DBLP:journals/corr/abs-1710-01013}. There is, however, a key difference between the SRNN and a feedforward network -- the iterative multiplication in the unrolled recurrent network is performed using the same weight matrix $\textbf{W}_{hh}$. In contrast, the multiplication in a feedforward network involves different weight matrices at each step ($\textbf{W}_1, \textbf{W}_2, \dots, \textbf{W}_{\ell_\text{max+1}}$). It can be argued that because the multiplication is carried out with different matrices, the fluctuations of the gradient magnitudes cancel each other out, thus reducing the possibility of the gradient going to zero. In reality, this balance is difficult to achieve. Glorot \& Bengio~\cite{Glorot10understandingthe} give an analytical explanation on why gradients in deep feedforward network can still vanish. Assume the network operates in a linear regime and the biases are set to zero. Let all weights be initialised independently, input feature variances be the same, and \textbf{x} and \textbf{W} be uncorrelated and have zero mean. If $n$ is the number of neurons in the input layer, the variance of $\boldsymbol{v}$ is given by

\begin{equation}
  \begin{aligned}
	& \text{Var}[v_1] = n \text{Var}[W_1] \text{Var}[x] \\
	& \text{Var}[v_{\ell}] = \text{Var}[x] \prod_{i=1}^{\ell} n_i \text{Var}[W_i] 
  \end{aligned}
\end{equation}

\noindent where $n_i$ is the size of layer $i$ and $\text{Var}[W_i]$ is the shared scalar variance of the weights in layer $i$. Let the network operate using a symmetric activation function chosen so that $\sigma'(0)=1$. As \textbf{x} and \textbf{W} are independent of each other and have zero mean $\sigma'(v_i) \approx 1, v_i \in \boldsymbol{v}_{\ell}$. Applying the delta method yields

\begin{equation}\label{eq:var-sigma}
  \begin{aligned}
	\text{Var}[\sigma(v_\ell)] & = \sigma'(E[v_\ell])^2 \text{ Var}[v_\ell] = \text{Var}[x] \prod_{i=1}^{\ell} n_i \text{Var}[W_i] 
  \end{aligned}
\end{equation}

Following \eqref{eq:ff-define}, the gradients used in backpropagation for layer $\ell$ are given by

\begin{flalign}
	& \frac{\partial \mathcal{E}}{\boldsymbol{v}_\ell} = \sigma'(\boldsymbol{v}_{\ell}) \textbf{W}_{\ell+1} \odot \frac{\partial \mathcal{E}}{\partial \boldsymbol{v}_{\ell+1}} \label{eq:dedvl}\\
	& \frac{\partial \mathcal{E}}{\partial \textbf{W}_\ell} = \textbf{h}_{\ell-1} \frac{\partial \mathcal{E}}{\partial \boldsymbol{v}_\ell} = \sigma(\boldsymbol{v}_{\ell-1})\frac{\partial \mathcal{E}}{\partial \boldsymbol{v}_\ell}
\end{flalign}

Using $\sigma'(v_i) \approx 1$ and \eqref{eq:dedvl}, for a network with $\ell_{max}$ hidden layers we get

\begin{equation}\label{eq:vardedvl}
	\text{Var}\bigg[\frac{\partial \mathcal{E}}{\boldsymbol{v}_\ell}\bigg] = \text{Var}\bigg[\frac{\partial \mathcal{E}}{\boldsymbol{v}_{\ell_{max+1}}}\bigg] \prod_{i = \ell}^{\ell_{max}} n_i \text{Var}[W_i]
\end{equation}

Using \eqref{eq:var-sigma} and \eqref{eq:vardedvl} in \eqref{eq:dedvl} gives

\begin{equation}
\begin{split}
	\text{Var}\bigg[\frac{\partial \mathcal{E}}{\boldsymbol{W}_\ell}\bigg] = \text{Var}\bigg[\frac{\partial \mathcal{E}}{\boldsymbol{v}_{\ell_{max+1}}}\bigg] \text{Var}[x] \\ \prod_{i=1}^{\ell-1} n_i \text{Var}[W_i]  \prod_{i = \ell}^{\ell_{max}} n_i \text{Var}[W_i]
\end{split}
\end{equation}

According to Glorot \& Bengio~\cite{Glorot10understandingthe}, this leads to the following two properties, provided the network layers have identical number of neurons and initialisation:

\begin{flalign}
	& \forall \ell, \text{Var}\bigg[\frac{\partial \mathcal{E}}{\boldsymbol{v}_\ell}\bigg] = \bigg[ n \text{Var}[W] \bigg]^{(\ell_{max} + 1 - \ell)} \text{Var}[x] \label{eq:property-1} \\
	& \forall \ell, \text{Var}\bigg[\frac{\partial \mathcal{E}}{\boldsymbol{W}_\ell}\bigg] = \bigg[ n \text{Var}[W] \bigg]^{(\ell_{max} + 1)} \text{Var}[x] \text{Var}\bigg[\frac{\partial \mathcal{E}}{\boldsymbol{v}_{\ell_{max+1}}}\bigg] \label{eq:property-2}
\end{flalign}

Equations \ref{eq:property-1} and \ref{eq:property-2} reveal that the variance of the backpropagated gradient might still vanish, despite the variance of the gradient of the weights being the same across the entire network. This has been experimentally confirmed by Bradley~\cite{Bradley-2010-10454}, who demonstrates that as the number of hidden layers between a selected layer and the output increases, the distribution of contributions from the selected layer peaks around zero.

\section{Solving gradient instability with learned orthogonality}
Different approaches have been attempted to alleviate the vanishing gradient problem. Long-Short Term Memory (LSTM) networks are a successful and widely used technique~\cite{Hochreiter1997}. They use a different architecture, replacing the hidden units with memory cells. Le et al.~\cite{Le2015}, however, demonstrate that standard recurrent neural networks with ReLU units and careful initialisation is capable of outperforming LSTMs on specific tasks. In the proposed configuration the weights are initialised using the identity matrix, an architecture Le et al. call IRNN. Arjovsky et al.~\cite{Arjovsky2015} propose recurrent neural networks with complex hidden units initialised with unitary matrices. Glorot \& Bengio~\cite{Glorot10understandingthe} propose \textit{normalized initialisation}, which tries to maintain back-propagated gradients variance by sampling from a Gaussian distribution within an interval that factors in the size of the network layers. The collective term \textit{initialisation tricks} is commonly used to describe this kind of careful initialisation techniques.

This paper looks specifically at \textit{orthogonal initialisation} in the context of biological plausibility. The basic idea of this type of initialisation is to use real orthogonal or semi-orthogonal matrices for the synaptic weights $\textbf{W}$. An important property of orthogonal matrices is that they are norm-preserving.

\begin{equation}
	||\textbf{W} \textbf{A}|| = ||\textbf{A}||, \forall \textbf{A}
\end{equation}

The analysis in Section~\ref{sec:vanishing_rnn} established that it is sufficient for $||\textbf{W}||_2 < 1$ for the gradients to vanish. Keeping \textbf{W} close to orthogonal ensures that the iterative multiplications preserve the norm of \textbf{W}, thus controlling the gain (i.e. spectral radius) of \textbf{W}. Setting $\textbf{W}_{hh}$ (for SRNN's) and $\textbf{W}_\ell$ (for feedforward networks) to random orthogonal matrices is an initialisation trick that has been widely used to provide optimisational advantage \cite{DBLP:journals/corr/HenaffSL16,DBLP:journals/corr/VorontsovTKP17}.

A common method for orthogonal initialisation is to start with random Gaussian matrices and compute QR factorisation to obtain an orthogonal matrix. It is, however, difficult to imagine such precise decomposition taking place in the brain. It can be argued that tricks like weight initialisation, conjugate gradients, and momentum have little obvious connection to neuroscience~\cite{Marblestone2016TowardsAI}.

On the other hand, there is physiological evidence of orthogonal structures in the primate and human brain. Using magnetic resonance imaging Wedeen et al.~\cite{Weeden2012} observed coherent and continuous grid structure of cerebral pathways in the three principal axes of embryonic development. Moreover, cortico-cortical pathways that form parallel sheets of interwoven paths were identified, with the paths within each sheet being orthogonal. Another example is the hippocampus, which has a broad functional role of memory-associated information processing \cite{PMID:10050854}. Gloveli et al.~\cite{Gloveli2005} provide physiological evidence for orthogonal relationship between behaviour specific gamma and theta oscillations in the hippocampus. This orthogonal arrangement of the microcircuits responsible for rhythm generation is present in the arborization patterns of internuncial neurons in the CA3 hippocampal area. Theta-band oscillations are observed predominantly along the longitudinal axis of the hippocampus, while gamma frequencies dominate the transverse axis. Gloveli et al.~\cite{Gloveli2005} suggest that morphological features like the innervation of the distal dendrites of the CA3 pyramidal cells underline the orthogonal arrangement. 

In light of the evidence above, it appears that orthogonal structures in the brain are not intrinsically implausible, although the objections against the techniques employed in the process of orthogonalisation still stand. Algorithmic approaches like the Gram-Schmidt process are indeed difficult to explain using mechanisms of the nervous system. This section of the paper demonstrates that it is possible to design an ANN that starts with random synaptic weights and evolves them to orthogonal matrices in the process of training. Two alternative options are presented: one that introduces a pre-training phase to the ANN, and another that naturally evolves the weights using constrained optimisation. Note, that the benefits of using soft orthogonal constraints have been well studied in the context of convolutional neural networks. Xie et al.~\cite{DBLP:journals/corr/XieXP17} develop a scheme where orthonormal constrain combined with batch normalisation and ReLU activations achieves 3\% $\sim$  4\% improvement for a 44-layer network on CIFAR-10. Bansal et al.~\cite{DBLP:journals/corr/abs-1810-09102} present a number of orthogonality regularisers and demonstrate consistent performance gains on popular datasets like CIFAR-10, CIFAR-100, SVHN and ImageNet. These two works can be viewed as a special case of penalty enforced orthogonality as such techniques are not limited by the network architecture. Indeed, we demonstrate that the weight matrices can be evolved towards orthogonality in both recurrent and feedforward networks (Section~\ref{sec:experiments}).

\subsection{Layer-wise orthogonal initialisation}\label{sec:layer-wise}
The first approach to imposing orthogonality applies a layer-wise pre-processing before the actual supervised training commences. This method considers the initialisation to be separate from the actual training, and envisions it more as a part of the initial development of the network. This can be viewed as a mechanism  of synaptogenesis responsible for forming orthogonal pathways in the brain. A related idea of initialising Boltzmann machines and Hopfield networks with a feedforward computation is presented in Bengio et al.~\cite{DBLP:journals/corr/BengioSBSS16}, where the authors also argue that the architecture is plausible from a biological standpoint.

During pre-training each layer $\ell$ performs local optimisation, minimising a pre-training cost function $\mathcal{E}^{p}_\ell$ that expresses the difference between $\textbf{W}_\ell \cdot \textbf{W}_\ell^T$ and the identity matrix. A suitable function can be the mean squared error (MSE), or the Frobenius norm

\begin{equation}\label{eq:min-norm}
	\mathcal{E}^{p}_\ell = \norm{(\textbf{W}_\ell \cdot \textbf{W}^{T}_\ell - \textbf{I})}^2_{F}
\end{equation}

The synaptic weights in each hidden layer are locally corrected using the following update rule

\begin{equation}
	\textbf{W}_\ell := \textbf{W}_\ell - \alpha_p \frac{\partial \mathcal{E}^{p}_\ell}{\partial \textbf{W}_\ell} 
\end{equation}

\noindent where $\alpha_p$ is a pre-training learning rate. The update process is performed sequentially or in parallel, until a certain convergence criterion is met. The criterion could be based on the magnitude of $\mathcal{E}^{p}$, the size of $\frac{\partial \mathcal{E}^{p}}{\textbf{W}}$, or the number of training iterations. Initially the synapses of the network are multiply innervated (fully connected using random initialisation). As the synapses mature, they segregate and undergo a form of synaptic elimination where the synaptic strength is modified on the basis of credit assignment. The $\mathcal{E}^p$ minimisation eliminates synaptic connections by pushing some elements of $\textbf{W}$ closer to its mean. It also strengthens other connections by pushing them away from the mean. The elimination process can be considered apoptotic \cite{doi:10.1111/j.1460-9568.2005.04377.x}, as this phase is a form of pre-training that is not related to the actual problem the network will be trained to solve.

In recurrent neural networks, where the same transition matrix is applied at each time step, the optimisation is performed with respect to $\textbf{W}_{hh}$:

\begin{equation}
	\textbf{W}_{hh} := \textbf{W}_{hh} - \alpha_p \frac{\partial \norm{(\textbf{W}_{hh} \cdot \textbf{W}_{hh}^{T} - \textbf{I})}^2_{F}}{\partial \textbf{W}_{hh}} 
\end{equation}

By further exploring \eqref{eq:min-norm} it can be shown that the minimisation of this function always leads to a high quality local minima (i.e. minima that are close to the global minimum of the energy landscape). Let $\textbf{X} \in \mathbb{R}^{m \times m}$ be a nonsingular matrix of independent and identically distributed random variables with mean $\mu_{x}$ and variance $\sigma_{x}^2$. Let $f(\textbf{X})$ be defined as

  \begin{equation}\label{eq:loss}
		f(\textbf{X}) =  \norm{(\textbf{X} \cdot \textbf{X}^{T} - \textbf{I})}^2_{F}
  \end{equation}

\noindent where $\textbf{I}$ is an $m \times m$ identity matrix, $\norm{\cdot}_{F}$ is the Frobenius norm, and $x_{ij}$ is the $(i,j)^\text{th}$ entry of $\textbf{X} \cdot \textbf{X}^{T}$ so that

 \begin{equation}
 \begin{split}
		f(\textbf{X}) &= \Bigg( \sqrt{\sum\limits_{i=1}^{m}\sum\limits_{j=1}^{m}|x_{ij}-\delta_{ij}|^2} \Bigg)^2 \\
		&= \sum\limits_{i=1}^{m}\sum\limits_{j=1}^{m}(x_{ij}-\delta_{ij})^2 , \{ \delta_{ii} = 1, \delta_{ij} = 0, i \neq j \}
  \end{split}
  \end{equation}

Expanding the above expression gives

\begin{equation}
\begin{split}
\sum\limits_{i=1}^{m}\sum\limits_{j=1}^{m}(x_{ij}-\delta_{ij})^2 & = \sum\limits_{i=1}^{m}\sum\limits_{j=1}^{m} (x_{ij}^2 - 2x_{ij}\delta_{ij}+\delta_{ij}^2) \\
 & = \sum\limits_{i=1}^{m} \Big( \sum\limits_{j=1}^{m} x_{ij}^2 - 2x_{ii} + 1\Big) \\
 & = \sum\limits_{i=1}^{m} \sum\limits_{j=1}^{m} x_{ij}^2 - 2\sum\limits_{i=1}^{m}x_{ii} + m
\end{split}
\end{equation}

Let $U = \sum\limits_{i=1}^{m} \sum\limits_{j=1}^{m} x_{ij}^2$  and $V = \sum\limits_{i=1}^{m}x_{ii}$. Then $f(\textbf{X}) = U - 2V + m$, where $U$ and $V$ are random variables and $m$ is a number. By definition $x_{ij}$ are independent and identically distributed with $E[\textbf{X}] = \mu_{x}$ and $Var(\textbf{X}) = \sigma_{x}^2$. According to the central limit theorem $U$ and $V$ tend toward a normal distribution and

\begin{equation}
\begin{split}
	E[U] & = m^2 (\sigma_{x}^2 + \mu_{x}^2) \\
	E[V] & = m \mu_{x}
\end{split}
\end{equation}

Moreover, since $f(\textbf{X}) = U - 2V + m$ is a sum of i.i.d. terms, $f(\textbf{X})$ also tends towards a Gaussian and

\begin{equation}
\begin{split}
	E[f(\textbf{X})] & = m^2 (\sigma_{x}^2 + \mu_{x}^2) - 2 m \mu_{x} + m \\
	Var(U) & = m^2 \; Var(\textbf{X}^2)\\
	Var(V) & = m  \sigma_{x}^2
\end{split}
\end{equation}
 
The variance of $f(\textbf{X})$ is then

\begin{equation}
\begin{split}
	Var(f(\textbf{X})) & = m^2 \; Var(\textbf{X}^2) + 2 m \sigma_{x}^2\\
					   & = m^2 \; \big( E[\textbf{X}^4] - E[\textbf{X}^2]^2\big) + 2 m \sigma_{x}^2 \\
					   & = m^2 \; E[\textbf{X}^4] - m^2 ( \sigma_{x}^2 + \mu_{x}^2)^2 + 2 m \sigma_{x}^2
\end{split}
\end{equation}
 
In the case when $\mu_{x} = 0$ we get
 
\begin{equation}\label{eq:variance}
	Var(f(\textbf{X})) = m^2 \; E[\textbf{X}^4] - m^2 \sigma_{x}^4 + 2 m \sigma_{x}^2
\end{equation} 

Let's look at the case where $\textbf{X}$ is sampled from a normal distribution -- $\textbf{X} \sim \mathcal{N}(\mu_x, \sigma^2_x)$. The fourth central moment of $\textbf{X}$ will be $E[\textbf{X}^4] = 3\sigma_{x}^4$, hence

\begin{equation}
    Var(f(\textbf{X})) = 2 m^2 \sigma_{x}^4 + 2 m \sigma_{x}^2
\end{equation} 

The fact that $f(\textbf{X})$ tends toward a normal distribution also holds when $\textbf{X} \sim \mathcal{U}(a,b)$. In this case the fourth moment is given by $E[\textbf{X}^4] = \frac{1}{5} \sum\limits_{i=0}^{4} a^i b^{(4-i)}$. More specifically, if $\textbf{X}$ is sampled from a uniform distribution that is symmetric around 0 ($\mathcal{U}(-b,b)$) we have

\begin{equation}
\begin{split}
    & E[\textbf{X}] = 0 \\
    & E[\textbf{X}^4] = \frac{3}{5} b^4 \\
    & Var(\textbf{X}) = \frac{1}{3} b^2
\end{split}
\end{equation}

Using \eqref{eq:variance} we get

\begin{equation}
\begin{split}
    Var(f(\textbf{X})) &= m^2 \; E[\textbf{X}^4] - m^2 \sigma_{x}^4 + 2 m \sigma_{x}^2\\
                       &= m^2 \frac{3}{5} b^4 - m^2 (\frac{1}{3} b^2)^2 + 2m\frac{1}{3} b^2\\
                       &= m^2 \frac{22}{45} b^4 + m\frac{2}{3} b^2
\end{split} 
\end{equation}

Now let's consider $f(\textbf{X})$ as the loss function of a high dimensional ($m \times m$) optimisation problem. The optimisation of $f(\textbf{X})$ over $\textbf{X} \in \mathbb{R}^{m \times m}$ can be viewed as a time continuous stochastic process $\{\textbf{X}_t; t \in T\}$ where $\textbf{X}_{t+1} := \textbf{X}_t - \alpha \nabla f(\textbf{X}_t)$. Assuming that the process is run for $N$ steps, the values for each time step can be gathered in a vector $\textbf{f} = (f(\textbf{X}_1),f(\textbf{X}_2), \cdots, f(\textbf{X}_N))^T$. As $f(\textbf{X})$ tends toward a normal distribution it is also true that $\textbf{f} \to \mathcal{N}(\mu_f, \Sigma)$. The optimisation problem can then be viewed as a one dimensional Gaussian Process $p(f(\textbf{X})) \sim \text{GP}(\mu,k(r))$ where $\mu = \mu_x$ and $k(r)$ is a squared exponential covariance function defined as $k(r) = \exp ^ {-\frac{1}{2}r^2}, r = \norm{\textbf{X} - \textbf{X}'}$.

The behaviour of gradient descent in Gaussian random fields (GRF) has been well studied~\cite{2018arXiv180309119C}. It has been shown that in this case local minima are located in a band close to the global minimum and that stochastic gradient descent (SGD) converges to high quality local minima, lower-bounded  by  the global minimum \cite{pmlr-v38-choromanska15}. As the minimisation of the loss given in \eqref{eq:loss} represents SGD in a one dimensional GRF, it is natural to conclude that it will always converge to a high quality minima, as long as $m$ is sufficiently large. Note, that this conclusion is not limited to the loss function given in \eqref{eq:loss}, but holds for any function that is a sum of i.i.d. terms. For example, instead of the square one could minimise the Frobenius norm directly, which should only impact the speed of convergence to the limit.

A simple experiment was performed using a $100 \times 100$ matrix sampled from a zero-mean Gaussian distribution with a standard deviation of 0.1 that was orthogonalised using the aforementioned procedure. The experiment was repeated 10,000 times using a value of $\alpha_p = 0.1$ and the convergence criterion was chosen to be $\mathcal{E}^{p} < 10^{-6}$. The success rate of the experiment was 100\% with an average convergence speed of only $\approx 22.77$ steps. Sampling the matrix from $\mathcal{U}[-0.1, 0.1]$ instead produces identical results with an average convergence speed of $\approx 24.00$.

\subsection{Penalty enforced orthogonality}\label{sec:penalty}
Another approach for developing and maintaining orthogonality is based on adding a penalty term that reduces the distance between $\textbf{W} \cdot \textbf{W}^T$ and the identity matrix. This constraint can take the form of

\begin{equation}
	\lambda \times  \norm{(\textbf{W} \cdot \textbf{W}^{T} - \textbf{I})}^2_{F}
\end{equation}

\noindent where $\lambda$ is a coefficient that governs the amount of regularization. This term can be added to the global network error and for classification tasks results in

\begin{equation}
	\mathcal{E} = -\frac{1}{N} \sum\limits_{i = 1}^{N} \textbf{t}_i \text{ log }(\textbf{y}_i) + \lambda \times  \norm{(\textbf{W} \cdot \textbf{W}^{T} - \textbf{I})}^2_{F}
\end{equation}

For SRNNs the penalty can specifically target the transition matrix by leaving the cost function intact and modifying the $\textbf{W}_{hh}$ update rule instead 

\begin{equation}
\begin{split}
	\textbf{W}_{hh_{t+1}} := \textbf{W}_{hh_{t}} - \alpha  \Bigl( \frac{\partial \mathcal{E}}{\partial \boldsymbol{W}_t} + 4 \times \lambda \times  \textbf{W}_{hh_{t}} \\ \times (\textbf{W}_{hh_{t}} \cdot \textbf{W}_{hh_{t}}^{T} - \textbf{I}) \Bigr)
\end{split}
\end{equation}

\section{Experiments}\label{sec:experiments}

A series of experiments\footnote{Full source code is made available at \url{https://github.com/nmanchev/OO-INIT-ANN}} were performed to validate the orthogonalisation techniques presented in the previous section. A feedforward network was tried on an MNIST classification task, and an SRNN was evaluated against two types of classification problems -- a set of synthetic problems and a sequential variant of the MNIST classification task.

\subsection{Pathological synthetic problems}

The pathological synthetic problems is a set of challenges originally outlined by Hochreiter \& Schmidhuber \cite{Hochreiter1997}. They were purposefully designed to be very difficult for a recurrent neural network to solve. Four of the original problems were selected -- the Temporal Order Problem, the 3-bit Temporal Order Problem, the Adding Problem, and the Random Permutation Problem.

Three independent SRNNs were trained on the problems listed above - a vanilla SRNN, an SRNN that uses the orthogonal initialisation suggested in Section~\ref{sec:layer-wise} (SRNN+OINIT), and an SRNN with the penalty enforcement scheme suggested in Section~\ref{sec:penalty} (SRNN+OPEN). Note that the SRNN network is identical to the SRNN+OPEN network with $\lambda=0$, and if the SRNN+OINIT's pre-training is omitted, then the SRNN+OINIT reduces to the vanilla SRNN.

The three networks were trained on each of the synthetic problems, starting with a sequence length of $T=10$. If a network manages to meet the success criterion of a problem for the specified sequence length, $T$ is increased by 10 and the network is trained and tested again. This cycle is repeated until the network fails to meet the success criterion. This process was employed to determine the maximum $T$ for all problems that each of the networks can handle.

The architecture was kept consistent across the networks. They were all configured to use a hyperbolic tangent activation function and 100 neurons in the hidden layer. The number of elements of $\textbf{W}_{xh}$ and $\textbf{W}_{hy}$ was allocated based on the specific problem. The parameter optimisation was performed using SGD. It has been shown that a simple initialisation of the synaptic weights by sampling from $\mathcal{N}(\mu=0,\sigma=0.1)$ leads to a network that is unable to solve the problems even for sequences as low as 10 \cite{ManchevSpratling2019}. Therefore, in order to provide a more meaningful measure of the impact of orthogonalisation, all synaptic weights were initialised using normalised initialisation \cite{Glorot10understandingthe}.

\begin{table*}[tpb]
\caption{Maximal depth (sequence length) by model}
\label{tab:depth}
\begin{center}
\begin{tabular}{lllllll}
Problem & \multicolumn{2}{c}{SRNN} & \multicolumn{2}{c}{SRNN+OPEN} & \multicolumn{2}{c}{SRNN+OINIT} \\ 
• & $T$ & parameters & $T$ & parameters & $T$ & parameters \\ 
\hline 
Temporal Order& 50 & $\alpha=0.01$ & 80 & $\alpha=0.001$,  $\lambda=1.0$ & \textbf{120} & $\alpha=0.0001$\\ 
3-bit Temporal Order & 50 & $\alpha=0.1$ & 70 & $\alpha=0.001$,  $\lambda=1.0$ & \textbf{90} & $\alpha=0.0001$\\ 
Adding & 80 & $\alpha=0.01$ & 80 & $\alpha=0.01$,  $\lambda=0.0001$ & \textbf{100} & $\alpha=0.01$\\ 
Random Permutation & 90 & $\alpha=0.0001$ & 140 & $\alpha=0.1$,  $\lambda=0.01$ & \textbf{240} & $\alpha=0.1$ \\ 
\hline 
\end{tabular} 
\end{center}
\end{table*}

The original success criteria given in \cite{Hochreiter1997} were used to consider a problem as solved for a given sequence length. For the classification problems (temporal order problem, the 3-bit temporal order problem, and the permutation problem) the network error uses a \textbf{softmax} function in its final layer and the global error is measured as the cross-entropy between prediction and a target. For the adding problem, where the output of the network is a continuous real value, the final layer of the network is linear and the global error is calculated using MSE.

For all problems the accuracy of the network is measured every 100 iterations on a test set of 10,000 samples. For the adding problem, a prediction is considered successful if the absolute difference between output and target is below $0.0001$. The learning for all problems was restricted to 100,000 iterations (one mini-batch of 20 training samples per iteration or $2\times10^6$ samples in total).

The results for all four problems and the maximum depth achieved by the networks are given in Table~\ref{tab:depth}. It is evident that introducing a penalty consistently leads to successful solutions at greater depth. Even so, using the orthogonal pre-training scheme outperforms both the SRNN and the SRNN+OPEN networks.

\begin{figure*}[htpb]
\begin{center}
\includegraphics[width=0.85\textwidth]{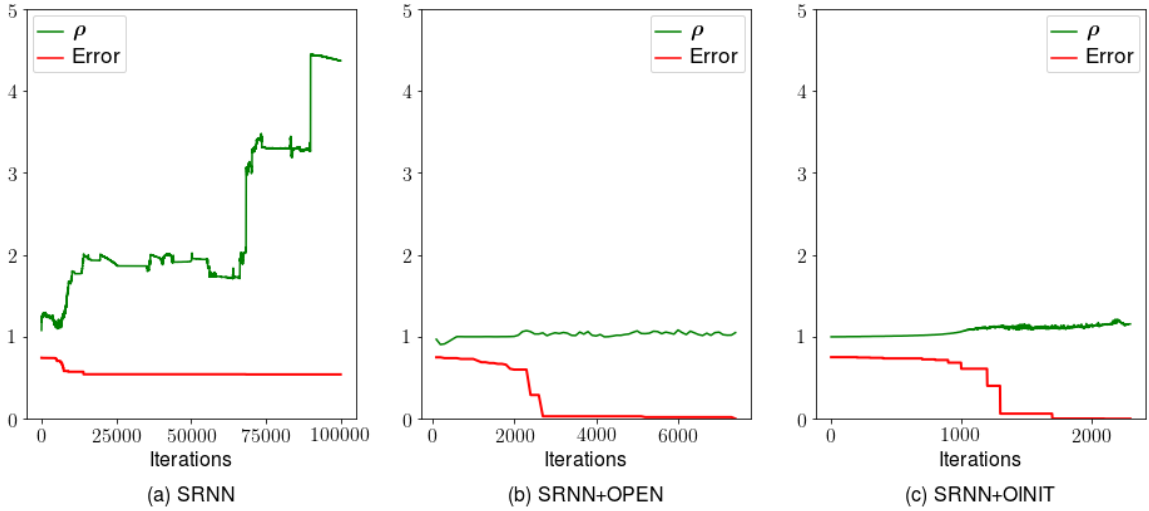}
\end{center}
\caption{Impact of penalty enforced orthogonality and orthogonal initialisation on solving the Temporal Order Problem for $T=60$.}
\label{fig:penalty1}
\end{figure*}

Figure~\ref{fig:penalty1} provides an intuition on why introducing a penalty term or orthogonal initialisation leads to better results. It is straightforward to establish that the spectral radius of the transition matrix of the vanilla SRNN network continuously increases as training progresses, fulfilling the necessary condition for exploding gradients ($\rho(\textbf{W}_{hh}) > 1.0$) and preventing effective learning. In contrast, adding a penalty term facilitates the maintenance of a stable gradient norm and allows the network to reach convergence. Likewise, the orthogonal initialisation helps with stabilising $\rho(\textbf{W}_{hh})$ and reduces the rate of increase of the norm, allowing the network to converge before any instability is encountered.

\subsection{MNIST sequence of pixels}
This challenge uses a dataset assembled by LeCun et al.~\cite{Lecun98gradient-basedlearning} and derived from the NIST Special Database 19. The data comprises a collection of images of handwritten digits, each image having dimension of 28x28 pixels. The ``MNIST sequence of pixels'' task was originally suggested in Le et al.~\cite{Le2015}, and presents a classification problem where the image is sequentially fed to a recurrent network one pixel at a time. The maximal depth is set to $T=784$ (28x28 pixels) and the network predicts the class of the image at the final time step. This results in a very long range dependency problem that SRNNs struggle to solve. 

Two SRNNs were trained on this tasks in an attempt to assess the impact of using learned orthogonality. The first network (SRNN+OINIT) was configured to use layer-wise orthogonal initialisation with $\alpha_p = 0.1$ and pre-training convergence criterion $\mathcal{E}^{p} < 10^{-6}$. The second network (SRNN+OPEN) relies on penalty enforced orthogonality and its weights were randomly initialised from $\mathcal{N}(\mu=0,\sigma=0.001)$, which is identical to the initialisation used in Le et al.~\cite{Le2015}.

MNIST is originally split into a training and test sets containing 60,000 and 10,000 images respectively. Both networks adhere to this initial split and their generalisation is measured exclusively on the subset of test images. The networks use hyperbolic tangent activation and 100 neurons in their recurrent layer. The learning rate $\alpha$ was selected using a grid search over $\{10^{-3}$,$10^{-4}$,$10^{-5}$,$10^{-6}$,$10^{-7}\}$, which was subsequently extended to include two additional values for $\sigma$ (0.01 and 0.1), however this only worsened the performance across all $\alpha$ values. For the SRNN+OPEN network $\lambda$ was chosen from $\{1.0, 0.1, 0.01\}$. Identical to  \cite{Le2015}, the networks were trained for $10^6$ iterations ($16\times10^6$ samples) and the highest accuracy and respective parameters for each network are reported in Table~\ref{tab:mnist-seq}.

\begin{table}[bpt]
\caption{Performance (in terms of accuracy) of the SRNN+OINIT and SRRN+OPEN networks on the MNIST sequence of pixels classification task.}\label{tab:mnist-seq}
\begin{center}
\begin{tabular}{lcc}
Network & Accuracy (\%) & Parameters \\ 
\hline 
SRNN & 11.35& - \\
SRNN+OINIT & 86.81& $\alpha=0.00001, \alpha_p = 0.1, \sigma=0.01$\\
SRRN+OPEN  & \textbf{91.81} & $\alpha=0.0001, \lambda=0.1, \sigma=0.001$\\
\end{tabular} 
\end{center}
\end{table}

\begin{figure*}[!ht]
\begin{center}
\includegraphics[width=.87\textwidth]{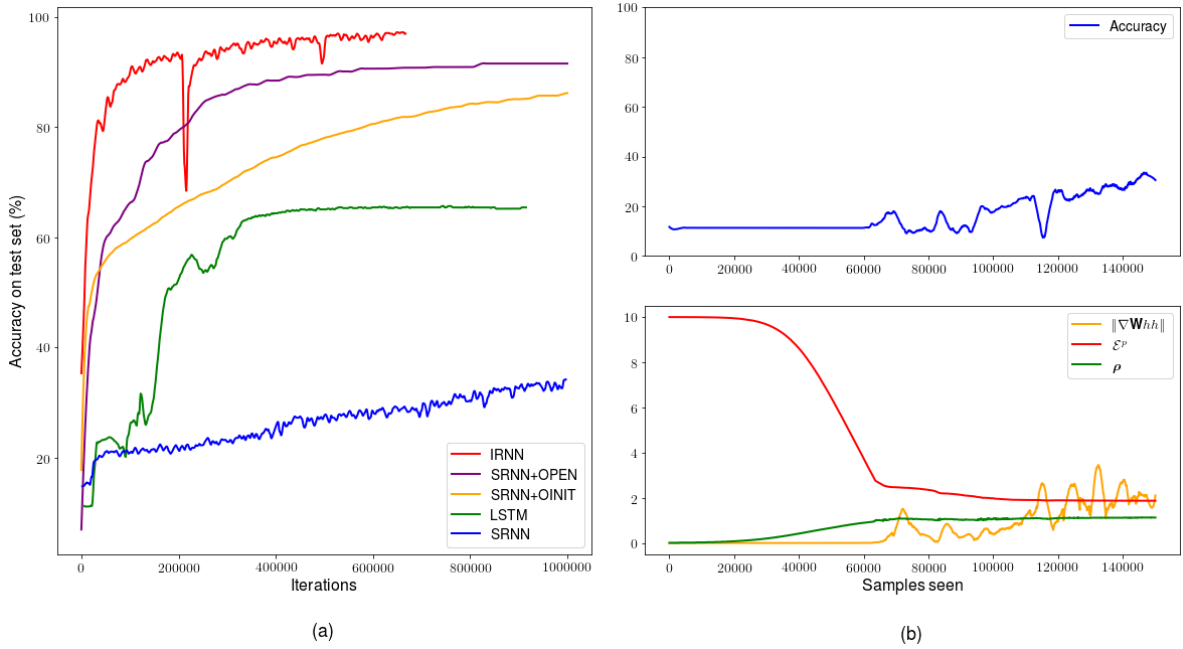}
\end{center}
\caption{\textbf{(a)} Accuracy of IRNN, LSTM, SRNN (data from \cite{Le2015}) and SRNN+OPEN and SRNN+OINIT networks on the MNIST  ``sequence of pixels'' task. \textbf{(b)} Learning dynamics of the SRNN+OINIT network}
\label{fig:mnist-sequence}
\end{figure*}

Figure~\ref{fig:mnist-sequence}a compares the SRNN+OINIT and SRNN+OPEN with the performance of an LSTM and  an SRNN networks as reported in \cite{Le2015}. The parameters of the LSTM and SRNN networks were also chosen on the basis of a grid search, and the parameters used for the networks in the plot are the ones that result in the best accuracy on the test set. The LSTM network has a learning rate of 0.01 and a forget gate bias of 1.0. The SRNN network has a learning rate of $10^{-8}$. Note, that the two networks also use gradient clipping \cite{Pascanu2012}, which mitigates the exploding gradients problem. To control for this, a separate test was conducted on an identical SRNN without gradient clipping. This network had an equivalent configuration and used the same grid search for selecting $\alpha$, however it failed to produce test accuracy above 11.35\%. This figure matches a baseline classifier that always predicts the mode of the test set ($1135/10000=11.35$), hence without clipping an SRNN network shows zero improvement with learning. In contrast the SRNN+OINT and SRNN+OPEN networks outperform the standard SRNN (with or without clipping) and the LSTM network, and their performance comes close to the IRNN network results given in \cite{Le2015}.

It is also worth looking at the early stages of the learning dynamics of the SRNN+OINIT network. Figure~\ref{fig:mnist-sequence}b depicts the change in the orthogonal penalty error term ($\mathcal{E}^p$), the spectral radius ($\rho$), and the gradient norm of the transition matrix. The values are tracked for the first 150,000 examples of the SRNN+OINIT network listed in Table~\ref{tab:mnist-seq}. At the start of the learning process the network suffers from vanishing gradients -- their norm and $\rho$ are settled near zero. As $\mathcal{E}^p$ decreases and $\textbf{W}_{hh}$ becomes closer to orthogonal, $\rho$ stabilises around 1.0, and the network begins to apply updates to the recurrent weights. This effects can also be observed by tracking the change in $\| \nabla \textbf{W}_{hh} \|$ and the global error measured on the test set. Whilst the network is in a vanishing gradient regime, the accuracy doesn't change and is settled at the baseline classifier value of approximately 11.35\%. The orthogonalisation of the synaptic weights allows the network to escape this regime after about 60,000 samples, and this is the same point in time when the reduction of global error begins.

\subsection{Feedforward MNIST task}

\begin{table*}[!htbp]
\parbox[t]{.45\linewidth}{
\caption{Best accuracy achieved on MNIST by the feedforward networks} \label{tab:mnist-ff}
\begin{center}
\begin{tabular}{lcc}
Network & Accuracy (\%) & Parameters \\ 
\hline 
FF & 11.35& -\\
FF+OPEN  & \textbf{97.03} & $\alpha=0.01, \lambda=0.01$\\
FF+OINIT  & 96.77 & $\alpha=0.01$\\
\end{tabular} 
\end{center}
}
\hfill
\parbox[t]{.45\linewidth}{
\caption{Impact of adding $\lambda=1.0$ penalty to the FF network}\label{tab:mnist-ff-open}
\begin{center}
\begin{tabular}{lcl}
Network & Accuracy (\%) & Parameters \\ 
\hline 
FF+OPEN  & 91.40 & $\alpha=0.1$\\
  & \textbf{91.65} & $\alpha=0.01$\\
  & 90.66 & $\alpha=0.001$\\
  & 88.80 & $\alpha=0.0001$\\
  & 72.81 & $\alpha=0.00001$\\
\end{tabular} 
\end{center}
}
\end{table*}

Section~\ref{sec:vanishing-ffnet} demonstrated that gradient instability is not constrained to recurrent networks. It is fairly straightforward to reproduce the problem in a feedforward network, which doesn't even have to be very deep. A simple feedforward network (FF) with 10 hidden layers was trained on the original MNIST problem. The network was configured with 784 inputs, 100 neurons in each hidden layer, and 10 neurons in the output layer. The global output and the output of the hidden layers was computed using Equation \ref{eq:ff-define} with a hyperbolic tangent activation function. All weights were sampled from $\mathcal{N}(\mu=0,\sigma=0.001)$ (as in \cite{Le2015}), and the network was trained for 100 epochs ($6 \times 10^6$ samples). $\alpha$ was selected from $\{10^{-1}$,$10^{-2}$,$10^{-3}$,$10^{-4}$,$10^{-5}\}$. Two modified versions of the network FF+OPEN and FF+OINIT were also trained and evaluated. For the FF+OPEN network $\lambda$ was selected from $\{1.0$, $0.1$, $0.01$, $0.001\}$. The pre-training learning rate for FF+OINIT was set to $\alpha_p = 0.1$ and the convergence criterion was chosen as $\mathcal{E}^{p} < 10^{-6}$. The best accuracy results for the networks and their respective hyperparameters are given in Table~\ref{tab:mnist-ff}.

The FF+OPEN and FF+OINIT networks significantly outperform the vanilla feedforward implementation. No best parameters for the FF network are listed as all attempted values for $\alpha$ failed to yield a performance above the baseline classifier. Adding an orthogonal penalty term to the FF network, however, enables it to escape the vanishing gradient regime for all values of $\alpha$ (see Table~\ref{tab:mnist-ff-open}). 

\section{Summary}

This paper demonstrates that orthogonal initialisation does not necessarily require a precise singular-value decomposition factorisation. Instead, the orthogonalisation can be naturally learnt as a form of pre-training, or alternatively -- the weights can gradually evolve as part of the training process. This could be considered a step towards biological-plausibility, as it shows that it is in principal possible to learn orthogonal weights rather than predefining them. In all experiments, the suggested orthogonalisation methods significantly outperform neural networks that use purely random initialisation. It was also analytically shown, that pre-training based on minimisation of a sum of i.i.d. terms will always converge, provided the network is sufficiently large. The same approach can be adopted for other non-trivial initialisation schemes (e.g. IRNN \cite{Arjovsky2015}).

A valid objection against the use of orthogonal matrices is that they impose restrictions not only on perpendicularity, but also on orthonormality (i.e. the rows of the matrix must be an orthonormal basis). A possible way of relaxing these constraints is to decompose the matrix of synaptic weights into multiple factors. For example, the distance from the synapse to the soma can be given by factor $d$. It has been shown that this distance impacts dendritic attenuation -- the farther the input is from the soma, the broader the resultant somatic postsynaptic potentials \cite{3681}. The size of the synaptic spine ($s$) could be another factor, giving an excitatory postsynaptic potential for synapse $i$ as $w_i = d_i \times s_i$. If a matrix of distances $\boldsymbol{D}$ is initialised at random and kept constant, and a matrix of synaptic sizes $\boldsymbol{S}$ is orthogonalised, their product will result in a synaptic matrix $\textbf{W}$ that represents pathways of varying length, but has the desired orthogonal property of preserving the gradients as only $\boldsymbol{S}$ is updated during the learning process. This idea is admittedly crude as it does not justify the orthonormality of $\boldsymbol{S}$, however, it suggests a possible direction for future research.

Another possible topic of interest is the combination of pre-training and orthogonal penalty. It is logical to assume that such combination would exhibit the desired properties of both regimes as it the network will start learning with its synaptic weights already in an orthogonal configuration, which will further be maintained stable by the penalty term. This investigation, however, is also left to future research.

\bibliographystyle{IEEEtran}
\bibliography{IEEEabrv,conference_101719}

\end{document}